\documentclass[letterpaper,runningheads]{belbi2016}

\def\journalissue{Text Analytics Conference, Adverse Drug Reactions Track, 2017.}
\def\paperidnum{}
\usepackage{url}
\usepackage{siunitx}
\usepackage{verbatim}
\usepackage{graphicx}
\title{Extracting adverse drug reactions and their context using sequence labelling ensembles in TAC2017}
\titlerunning{Extracting ADRs using sequence labelling ensembles in TAC2017}
\author{Maksim Belousov\inst{1} \and Nikola Milosevic\inst{1,4} \and William Dixon\inst{2,3} \and Goran Nenadic\inst{1,2}}
\authorrunning{Maksim Belousov et al.} 
\institute{School of Computer Science, University of Manchester, UK\\
\and
Health eResearch Centre, Farr Institute, Manchester Academic Health Science Centre, The University of Manchester, UK \\
\and 
Arthritis Research UK Centre for Epidemiology, The University of Manchester, UK \\
\and 
Manchester Institute of Innovation Research, Alliance Manchester Business School, The University of Manchester, UK \\
  \email{\{maksim.belousov,nikola.milosevic,will.dixon,gnenadic\}@manchester.ac.uk}
  }

\begin{document}

\maketitle

\begin{abstract}

Adverse drug reactions (ADRs) are unwanted or harmful effects experienced after the administration of a certain drug or a combination of drugs, presenting a challenge for drug development and drug administration. In this paper, we present a set of taggers for extracting adverse drug reactions and related entities, including factors, severity, negations, drug class and animal. The systems used a mix of rule-based, machine learning (CRF) and deep learning (BLSTM with word2vec embeddings) methodologies in order to annotate the data. The systems were submitted to adverse drug reaction shared task, organised during Text Analytics Conference in 2017 by National Institute for Standards and Technology, achieving F1-scores of 76.00 and 75.61 respectively. 

\vspace{6pt}\textbf{Keywords:} health informatics, text mining, drug labels, adverse drug reactions

\end{abstract}

\section{Introduction}
Adverse drug reactions (ADRs) are unwanted or harmful effects experienced after the administration of a certain drug or a combination of drugs~\cite{lee2006adverse}. They present a challenge for drug development and drug administration. During 1994, it was estimated that 700,000 patients in the United States suffered from adverse drug reaction, while 100,000 died as a consequence of such reactions~\cite{lazarou1998incidence}. Roughly half of the people in the UK take prescribed medications. Adverse drug reactions are serious burden on health care systems. About 7\% of all hospital admissions were accounted to ADRs. Moreover, quality of life and adherence to treatment is, as well, affected by adverse drug reactions~\cite{pirmohamed2004adverse}. Also, they are important source of human phenotypic data and can be used to predict drug targets~\cite{kuhn2015sider}.

In the United States, drug product labels are required by law to contain the information regarding clinically significant adverse drug reactions~\cite{us2014cfr}. 
All drug product labels in the United States are freely available through the National Library of Medicine's DailyMed website\footnote{\url{https://dailymed.nlm.nih.gov/dailymed/index.cfm}} in a standard format called Structured Product Label (SPL). 

The task of recognising specific mentions (such as ADRs) in a text is a task of named entity recognition (NER) or tagging, which can be approached using sequence labelling techniques. Sequence labelling problems are usually solved using sequence modelling machine learning techniques, such as hidden Markov models, conditional random fields or recurrent neural networks.

Within the drug informatics domain, the SPLICER system~\cite{duke2013consistency} was successfully applied to extract adverse drug events from text and tables in the Adverse Reactions section of SPLs. Other efforts focus on side effects and drug indications~\cite{fung2013extracting,khare2014labeledin,boyce2012using}. The SIDER (Side Effect Resource) database uses named entity recognition to extract side effects and indications from product labelling, including SPLs~\cite{kuhn2015sider}. More recently, starting with full-text papers from the Journal of Oncology,  drug side effect relationships were extracted and compared to the SIDER database~\cite{xu2015large}. 

Neural networks with word embeddings have recently showed successes in the biomedical named entity recognition. Word2vec embeddings with bidirectional recurrent neural networks combined with a CRF tagger and SVM classifier showed promising results for disease recognition \cite{wei2016disease}. Named entity recognition methodology based on recurrent neural networks and word embeddings (GloVe or Word2vec) was used for de-identification of electronic health records and gave the state-of-the-art results,  producing slightly better results with GloVe embeddings \cite{dernoncourt2017identification}.

In this paper, we present our approaches to the recognition of adverse drug reactions and related entities, developed for a shared task organised during the Text Analytics Conference 2017 (TAC 2017). The task was co-organised by the US National Institute of Standards and Technology (NIST) and the US Food and Drug Administration (FDA). The objective of the shared task was to extract adverse drug reactions from drug labelling text documents using natural language processing techniques. In the task 1, in which we participated, the participants were supposed to build a system to extract adverse drug reactions and related mentions such as severity, drug class, negation, factors, and whether it was reported on animals\footnote{\url{https://bionlp.nlm.nih.gov/tac2017adversereactions/}}.
\subsection{Data}
The shared task organisers published a training dataset containing 101 annotated drug labels (documents) and a dataset containing 2,208 unannotated drug labels. An unseen subset of unannotated documents was used as a testing data during the task evaluation. The drug label is a multi-section document that may contain headings, paragraphs, tables and lists. In the provided dataset each drug label was converted to a text document disregarding the structure (i.e. representing all elements as an unformatted text, keeping only the main sections of the document). It is worth noting that the gold-standard dataset contained some \emph{discontinuous annotations} (6.8\% of all annotations). Annotation that involves more than one continuous span of characters is considered discontinuous annotation. For the simplicity of tagging schemes, we ignored discontinuous annotations during the document parsing.

The class distribution of annotated entities is imbalanced, where the majority of annotations were adverse drug reactions. On the other hand, some related entities had only a few annotations. The numbers of annotated mentions (groups of tokens), tokens and the average number of tokens per mention are presented in Table~\ref{table:data}. Lack of data for certain related entities presented a challenge for developing named entity recognition systems based on machine learning.
\begin{table*}[htbp]
\centering
\begin{tabular}{ lrrr }
  \hline
  \textbf{Entity class}  & \textbf{\#mentions} & \textbf{\#tokens} & \textbf{Avg. tk/mention}\\ \hline
  \small{Adverse drug reaction} & \small{12,792} & \small{21,258}  & \small{1.66} \\
  \small{Severity} & \small{863} & \small{1,306}& \small{1.51} \\
  \small{Factor} & \small{602} & \small{653}& \small{1.08} \\
  \small{Drug class} & \small{248}& \small{518}& \small{2.09} \\
  \small{Negation} & \small{95}& \small{109}& \small{1.47} \\
  \small{Animal} & \small{44} & \small{44}& \small{1.00} \\

\hline
\end{tabular} 
\caption{The number of annotated mentions (group of tokens), number of tokens, and the average number of tokens per mention in the provided training data}
\label{table:data}
\end{table*}

\section{System description}

\begin{figure}[h!]
\centering
\includegraphics[width=0.9\textwidth]{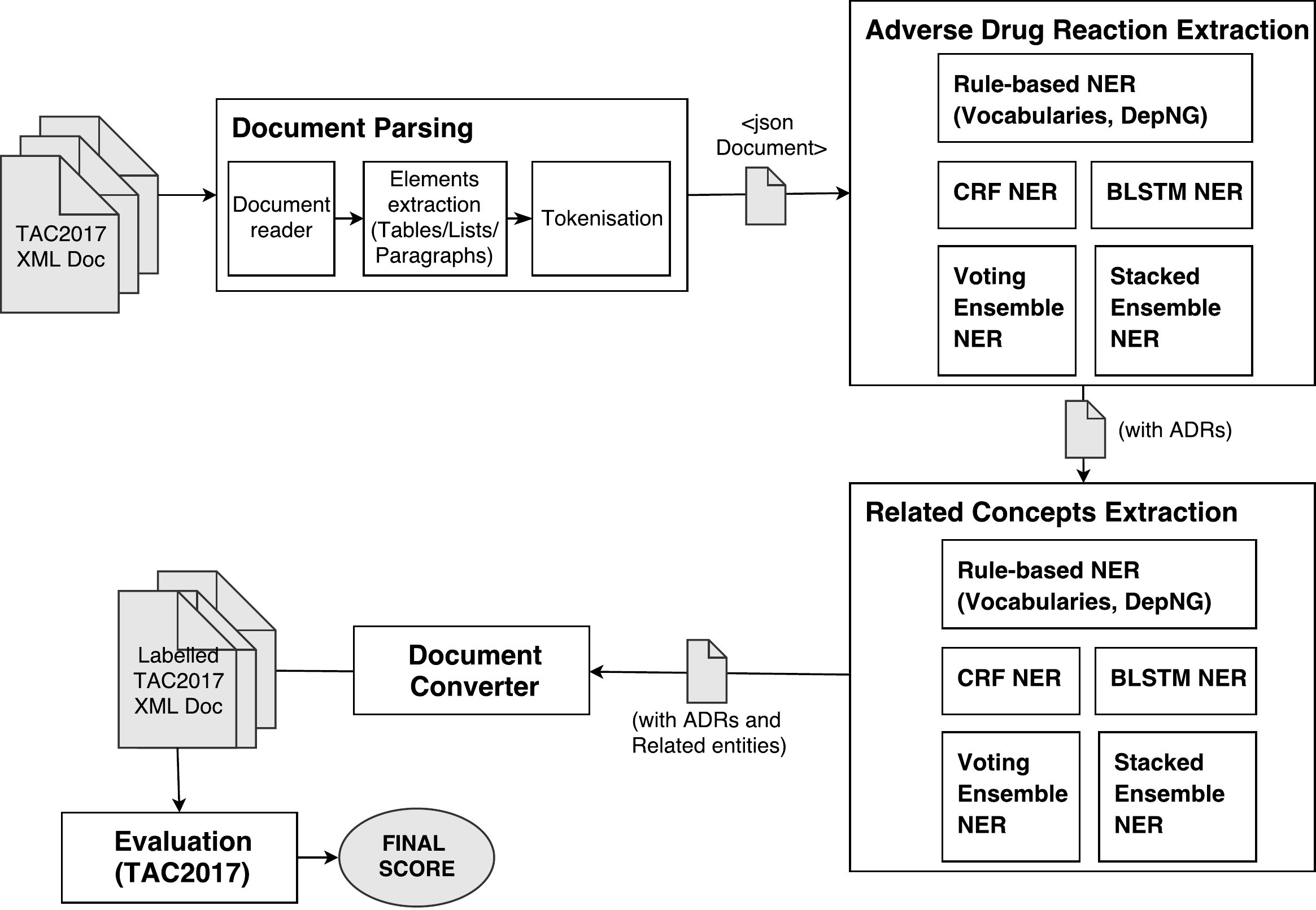}
\caption{Document processing and tagging pipeline.}
\label{fig:pipeline}
\end{figure}

The architecture of the proposed systems consists of three stages: (1) document parsing, (2) word vectorisation, (3) tagging ADRs and their related entities. During the document parsing stage we attempt to restore the original structure of the document and recognise elements such as headings, tables (with rows and cells), lists (with items) and text paragraphs. The word vectorisation stage depended on the type of tagging model in the following stage and aimed to generate word vectors from text sequences using either hand-crafted features or unsupervised learning. The main task of tagging stage is to extract mentions of specific type from text by sequence labelling of extracted word vectors. Since some related entities rely on ADR mentions, they are performed separately, after the ADR tagging is completed. The pipeline is presented in Figure~\ref{fig:pipeline} and the following subsections provide details about each processing stage.

\subsection{Document parsing}
The aim of this stage is to perform re-engineering the structure of the document so that later their content can be treated differently. For instance, it might be beneficial to analyse the content of a table cell individually rather than the whole chunk of the text that contain multiple rows and cells.
We identified four different element types in the document:
\begin{itemize}
	\item \textbf{Headings} are numbered titles for sections and sub-sections (e.g. \textit{``5.1     Asthma-Related Death [See Boxed Warning]"})
	\item \textbf{Tables} have \emph{heading rows} and \emph{content rows}, each of them is also having \emph{cells}. Each row might have different number of cells. In addition, a table may have \emph{caption} (which usually starts with ``Table NUM.'') and a \emph{footer} that contain additional notes. We treated all text lines after the aforementioned caption trigger and before the paragraph separator (multiple empty lines) as potential table rows. Then, we categorised each row candidate as part of the caption, header, content and footer, based on the number of potential columns, numerical cells and words in each cell.
	\item \textbf{Lists} are groups of multiple bullet-points or \emph{items}. Consequent text lines that starts with asterisk character (\textit{*}) are considered as list items. List should have more than one item.
	\item \textbf{Paragraphs} are any other chunk of text separated with multiple new-line characters.
\end{itemize}

For some tagging models we applied two different document \emph{splitting strategies}: (1) take the \textit{\textbf{whole element}} (i.e. table, list, paragraph) and represent them as text or (2) take the textual content of \textit{\textbf{sub-elements}} (such as table cells and list items) and treat them as individual items.

\subsection{Tagging models}
We utilised various types of tagging methods based on knowledge-driven rules, conditional random fields (CRF), bidirectional long short-term memory networks (BLSTM) and two different types of ensemble methods.   
We generated word vectors differently depending on the sequence labelling approach by using either hand-crafted features or obtaining word embeddings from unsupervised learning models trained on large text corpora.

\subsubsection{Rule-based models}
\paragraph{Rule-based} methods are based on a knowledge-driven approach and manually curated dictionaries. In particular, we applied them for \emph{negation} and \emph{animal} classes, since there was not enough labelled data to be modelled by machine learning algorithms.

\begin{itemize}
	\item To identify \textbf{negations}, we have developed a rule based tagger using the modification of DepND\footnote{\url{https://github.com/zachguo/DepND}} that uses GENIA dependency parser~\cite{sagae2007dependency} to recognise the scope of the negation and the dictionary of negation triggers. In particular, we added a list of phrases that need to be ignored if appeared in a negation phrase or scope (such as ``not available" or ``could not be assessed") and labelled negations only when an ADR mention is found inside the negation scope. We applied the negation tagger on the sub-element level (i.e. on table cells and list items).
	\item For the \textbf{animal} class, we made an assumption that animals are not mentioned in drug labels unless adverse events are reported on them during the trials. Also, there is a close set of animal spices that are usually used in medical experiments~\cite{mukerjee1997trends}. We have developed a dictionary-based tagger that labelled all mentions of animals from our list. The animal tagger was used on the sub-element level.
\end{itemize}

\subsubsection{CRF models}

\paragraph{Linear chain conditional random fields (CRF)} is a linear statistical model that encodes conditional distributions $p(y|x)$ between observations (input features) and output variables (labels). Prior to passing a text input into the model, each sequence item (i.e. word or token) should be converted into a \emph{feature vector}. In particular, we experimented with lexical features, part-of-speech tags, grammatical relations (dependencies), vocabulary and semantic features (such as corresponding semantic types and named entity tags from various medical systems). In order to capture the context for a given token, the mentioned features were extracted from a certain number of surrounding tokens (context window). All CRF models were used on the whole elements (i.e. tables, lists) represented as a text.

\begin{itemize}
	\item For \textbf{ADR} mentions, we extracted word lemmas, part-of-speech tags (retrieved using the GENIA tagger~\cite{tsuruoka2005developing}), UMLS semantic types (obtained using QuickUMLS~\cite{soldaini2016quickumls}) and lexicon match (i.e. whether the current word is exist in the ADR lexicon\footnote{\url{http://diego.asu.edu/downloads/publications/ADRMine/ADR_lexicon.tsv}}). We trained word2vec on lemmatised sentences extracted from 2,208 unannotated drug labels that were provided as a part of this task. In particular, we extracted 200-dimensional feature vectors from continuous bag-of-words model with a context window of size 5, trained with negative sampling using five noise words. Then we performed K-means clustering (n=50) of the word-vector space. For words that are found in the model, we used their corresponding cluster number, otherwise we used the lemma of the word as a feature. In order to capture the context we also extracted features from surrounding words (i.e. five preceding and five following words).
	\item For the \textbf{severity}, \textbf{factor} and \textbf{drug class}, we used a similar set of features with additional lexicon features. In particular, a lexicon for drug class was obtained from DrugBank\footnote{\url{https://www.drugbank.ca/}} and Anatomical Therapeutic Chemical Classification System (ATC)\footnote{\url{http://www.atccode.com/}}, whereas for other aforementioned classes we experimented with lexicons obtained from the provided labelled data. We also added an additional binary feature that indicates whether the ADR is mentioned in the surrounding context.
\end{itemize}

\subsubsection{BLSTM models}

\paragraph{Bidirectional Long Short-Term Memory networks (BLSTM)} are specific type of recurrent neural networks designed to learn long-term dependencies. In order to increase the amount of input information, the given sequence is read in both directions (forward and backward). For this tagging model we obtained word vectors from multiple word2vec models trained on large text corpora from generic and target domains. The generic 200-dimensional word embeddings were trained on a combination of PubMed and PMC texts with texts extracted from a recent English Wikipedia dump~\cite{moen2013distributional}, whereas the target 200-dimensional word embeddings were trained on 2,208 unannotated drug labels. The \emph{BLSTM} model  was trained using RMSprop~\cite{tieleman2012lecture} algorithm with the learning rate of \num{1e-5}. For regularisation, dropout with the rate of $0.1$ was applied on each LSTM layer with 170 units. We trained BLSTM model for 50 epochs and used early stopping with patience of 10 epochs. Since this model does not rely on hand-crafted features, we used the same model configuration for both adverse reactions and related entities. For all entity types, we have trained a single BLSTM model on the whole elements (i.e. tables, lists) represented as a text.

\subsubsection{Ensemble models}
We have created two different ensemble models:
\begin{itemize}
\item A \textit{voting BLSTM and CRF ensemble} was training both CRF and BLSTM classifiers in parallel and selected the best candidate based on the highest average predicted probability of each class obtained from each classifier.
\item A \textit{stacked CRF-BLSTM ensemble} is our proposed modification of Wolpert's stacked generalisation~\cite{wolpert1992stacked} that firstly trains the CRF classifier, using the previously described features, and then utilises its predicted probabilities for each class to build an additional token-level embeddings for the BLSTM classifier. In this way, the obtained word vector has the dimension of the number of target classes used in CRF and its values will correspond to predicted probabilities.
\end{itemize}

For the voting and stacked ensembles we have utilised an ADR-specific feature extractor and trained a single ensemble model on all classes.

 
\section{Evaluation of the tagging models on the training data}

The provided labelled data contained 101 documents. We evaluated the supervised machine learning models using holdout cross-validation; therefore the dataset was split into training (56 documents), validation (24 documents) and testing (21 documents) sets. The rule-based models were evaluated on the whole dataset.
The evaluation results for all developed taggers are presented in Table \ref{table:eval}.

\begin{table*}[htbp]
\centering
\begin{tabular}{ llrrr }
  \hline
  \textbf{Entity class}  & \textbf{Method} & \textbf{Precision} & \textbf{Recall} & \textbf{F1-score}\\ \hline
  \small{ADR} & \small{CRF} & \small{90} & \small{82} & \small{86} \\
  \small{} & \small{BLSTM} & \small{86} & \small{84} & \small{85}\\
  \small{} &\small{Voting BLSTM+CRF} & \small{91} & \small{84} & \textbf{\small{87}}\\
  \small{} &\small{Stacked CRF+BLSTM} & \small{90} & \small{85} & \textbf{\small{87}}\\ 
  \hline
  \small{Severity} &\small{CRF} & \small{67} & \small{51} & \small{58}\\
  \small{} &\small{BLSTM} & \small{55} & \small{75} & \small{64}\\ 
  \small{} &\small{Voting BLSTM+CRF} & \small{70} & \small{65} & \textbf{\small{67}}\\
    \small{} &\small{Stacked CRF+BLSTM} & \small{58} & \small{71} & \small{64}\\ \hline
  \small{Factor} &\small{CRF} & \small{52} & \small{20} & \small{29}\\
  \small{} &\small{BLSTM} & \small{73} & \small{46} & \small{56}\\ 
  \small{} &\small{Voting BLSTM+CRF} & \small{87} & \small{36} & \small{51}\\ 
    \small{} &\small{Stacked CRF+BLSTM} & \small{82} & \small{41} & \small{55}\\ \hline

  \small{Drug class} &\small{CRF} & \small{41} & \small{35} & \textbf{\small{38}}\\
  \small{} &\small{BLSTM} & \small{57} & \small{21} & \small{31}\\ 
  \small{} &\small{Voting BLSTM+CRF} & \small{62} & \small{12} & \small{20}\\
    \small{} &\small{Stacked CRF+BLSTM} & \small{57} & \small{24} & \small{34}\\ \hline

  \small{Negation} &\small{CRF} & \small{25} & \small{18} & \small{21}\\
  \small{} &\small{BLSTM} & \small{22} & \small{12} & \small{15}\\
    \small{} &\small{Voting BLSTM+CRF} & \small{50} & \small{06} & \small{11}\\
    \small{} &\small{Stacked CRF+BLSTM} & \small{57} & \small{24} & \small{33}\\ 
      \small{} &\small{Rule-based} & \small{66} & \small{66} & \textbf{\small{66}}\\ \hline

  \small{Animal} &\small{CRF} & \small{76} & \small{100} & \small{87}\\
  \small{} &\small{BLSTM} & \small{100} & \small{46} & \small{63}\\
      \small{} &\small{Voting BLSTM+CRF} & \small{100} & \small{38} & \small{56}\\
        \small{} &\small{Stacked CRF+BLSTM} & \small{40} & \small{31} & \small{35}\\
  \small{} &\small{Rule-based} & \small{86} & \small{100} & \textbf{\small{93}}\\

\hline
\end{tabular} 
\caption{Token-level evaluation of the taggers by the entity class and method used on the provided 101 labelled documents using holdout cross-validation.}
\label{table:eval}
\end{table*}

As it can be seen from Table \ref{table:eval}, we calculated precision, recall and F1-score for labelling tokens in the document. Later, sequential labels are post-processed and merged into mentions.

Both ensemble models usually outperformed individual models especially in cases where there was enough training and testing samples. The stacked and voting ensembles performed relatively similar, although the stacked ensemble was slightly better in general. The F1-score for labelling adverse drug reactions ranges between 85\%-87\%, with the maximum score for the ensemble and BLSTM tagger. The BLSTM tagger performed better on the severity and factor classes. Drug class gave the best results on the test set with the CRF tagger, however, these results were quite unstable. While CRF performed on the test set with F1-score of 38\%, on the validation set the F1-score was only 22\%. The rule based approach gave the best results for the rare classes, such as negation and animal.


\section{Runs and system evaluation}
Using the evaluation results presented in the previous section, we have combined the best-performing taggers and created two systems which correspond to the two runs submitted for the final shared task evaluation (on test data). 

\begin{itemize}
	\item \textbf{Run \#1:} We applied the \texttt{rule-based} approaches for the \emph{Negation} and \emph{Animal} classes. For \emph{Adverse Drug Reactions} we utilised the \texttt{CRF} model with  the hand-crafted features. For all other entity types (i.e. \emph{Severity}, \emph{Factor} and \emph{Drug class}) we used the BLSTM tagger. The three related entities used one BLSTM model.
	\item \textbf{Run \#2:} The \texttt{rule-based} tagger was applied only for the \emph{Negation} class, whereas all other classes were handled with the \texttt{Stacked CRF+BLSTM} ensemble model.
\end{itemize}
\subsection{Results}
The systems were trained on the whole annotated dataset provided (101 documents) and applied on unannotated dataset for automatic tagging (2,208 documents). Then, sample of the automatically tagged documents were used for the evaluation. The primary metric for this evaluation was the micro-averaged F1-score.
We have presented the system evaluation results in Table~\ref{table:submissions}.

\begin{table*}[htbp]
\centering
\begin{tabular}{ l|rrr|rrr }
  \hline
   \small{} & \multicolumn{3}{c|}{\textbf{Considering entity type}} & \multicolumn{3}{c}{\textbf{Not considering entity type}} \\
  \textbf{Submission} & \textbf{Precision} & \textbf{Recall} & \textbf{F1-score} & \textbf{Precision} & \textbf{Recall} & \textbf{F1-score}\\ \hline
  Run \#1 & 80.19 & 72.23 & \textbf{76.00} & 80.19 & 72.23 & 76.00 \\ 
  Run \#2 & 76.84 & 74.36 & \textbf{75.58} & 76.87 & 74.39 & 75.61 \\ 
\hline
\end{tabular} 
\caption{Performance of the submitted systems on the test data considering and not considering types of annotated entities. The primary metric used for the evaluation is marked in bold.}
\label{table:submissions}
\end{table*}

\subsection{Discussions}
The submitted systems had similar performance, with Run 1 having slightly better performance on the test data (by less than 0.5\%). The achieved results are similar to the results obtained on the training data using 3-fold cross-validation (F1-score of 77.26 for the Run 1, and 76.61 for the Run 2). 

The classes  were not balanced. Some classes, such adverse drug reactions had a fair number of labelled entities in the training set, and therefore the machine learning models could be efficiently trained on this class. However, other classes were relatively small compared to the ADR class. Also, other classes were related to the ADR class and were only triggered if the ADR class is labelled in its vicinity.  Context of the labels had significant importance in this task, as the same phrase is labelled depending on whether it is in vicinity of an ADR and whether it closer describes an ADR. For example word ``serious" will be labelled as severity in context of ``serious headache", however, it will not be labelled in other contexts, such as for example in ``serious consideration". 

On the other hand, some classes, such as animal and negation had only a small number of annotations in the training dataset. Therefore, it was impossible to train a machine learning model and it was necessary to develop a rule based approach. The rules for the negation class were considering context and whether in the scope of the negation is present an ADR. On the other hand, mentions of animals unrelated to an ADR were rare. Therefore, it was safe to make an assumption that all animal mentions are related to adverse drug reactions. 

\subsection{Conclusion}
In this paper we presented a number of different methodologies for labelling adverse drug reactions and related factors, severity, drug class, negation and whether they were reported on animals. We presented two systems made out of the best performing taggers that were submitted to the ADR track shared task of the Text Analytics Conference (TAC2017). The systems performed with F1-scores of 76\% and 75.58\% on the testing data.

There is still space for improvement of the system and performing additional experiments. More informative features of the text could help improve the CRF machine learning taggers, while more representative word embeddings could be helpful for the BLSTM based taggers. This can be achieved using additional vocabularies, semantic resources and knowledge bases.  

Other potential way to improve the performance of the tagging is to investigate alternative ensemble methods, e.g. utilise an additional meta-classifier to combine the CRF and BLSTM results.
In addition, performance of the BLSTM model directly depends on the word embeddings that were used, therefore alternative word representation models in addition to word2vec  might be utilised (e.g. multi-level word representation or knowledge-infused word embeddings).

However, there is still challenge of labelling classes that have a low number of examples. In these cases, it is challenging to create a good performing machine learning models, because of the lack of examples. However, our rule based approaches can be further improved with additional samples and looking at additional data. Also, machine learning performance can be probably improved by using additional annotated data and external data sets.  

\bibliographystyle{splncs03}
\bibliography{Literature}

\end{document}